\documentclass{article}

\PassOptionsToPackage{numbers, compress}{natbib}

\usepackage[final]{neurips_2023}
\usepackage{microtype}
\usepackage{graphicx}
\usepackage{caption}
\usepackage{subcaption}
\usepackage{booktabs} 

\usepackage{newfloat}
\usepackage{stfloats}

\makeatletter

\usepackage{natbib}
\usepackage[utf8]{inputenc} 
\usepackage[T1]{fontenc}    
\usepackage[hidelinks]{hyperref}       
\usepackage{url}            
\usepackage{booktabs}       
\usepackage{amsfonts}       
\usepackage{nicefrac}       
\usepackage{microtype}      
\usepackage{xcolor}         

\usepackage{amsmath}
\usepackage{amssymb}
\usepackage{mathtools}
\usepackage{amsthm}

\usepackage{MnSymbol}
\usepackage{tikz}
\usepackage{program}
\usepackage[all]{xy}
\usepackage[arrowdel]{physics}
\usepackage{multirow}
\usepackage[inline]{enumitem}

\usepackage{paralist, tabularx}

\usepackage[capitalize,noabbrev]{cleveref}

\theoremstyle{plain}
\newtheorem{theorem}{Theorem}[section]

\theoremstyle{definition}
\newtheorem{definition}[theorem]{Definition}

\theoremstyle{remark}

\newcommand*{\eg}{%
  {\textit{e.g.}}%
}

\newcommand*{\ie}{%
  {\textit{i.e.}}%
}

\usepackage{amsmath,amsfonts,bm}

\def\eqref#1{equation~\ref{#1}}

\def\1{\bm{1}}

\def\vzero{{\bm{0}}}
\def\vone{{\bm{1}}}

\def\vf{{\bm{f}}}
\def\vg{{\bm{g}}}

\def\vr{{\bm{r}}}

\def\vx{{\bm{x}}}

\def\mA{{\bm{A}}}

\def\mC{{\bm{C}}}
\def\mD{{\bm{D}}}

\def\mH{{\bm{H}}}
\def\mI{{\bm{I}}}

\def\mL{{\bm{L}}}

\def\mO{{\bm{O}}}

\def\mU{{\bm{U}}}

\def\mW{{\bm{W}}}
\def\mX{{\bm{X}}}

\def\mTheta{{\bm{\Theta}}}

\DeclareMathAlphabet{\mathsfit}{\encodingdefault}{\sfdefault}{m}{sl}
\SetMathAlphabet{\mathsfit}{bold}{\encodingdefault}{\sfdefault}{bx}{n}

\def\sH{{\mathbb{H}}}
\def\sI{{\mathbb{I}}}

\def\sO{{\mathbb{O}}}

\def\sR{{\mathbb{R}}}
\def\sS{{\mathbb{S}}}

\def\sV{{\mathbb{V}}}
\def\sW{{\mathbb{W}}}
\def\sX{{\mathbb{X}}}

\title{Fast Temporal Wavelet Graph Neural Networks}

\newcommand{\printfnsymbol}[1]{%
  \textsuperscript{\@fnsymbol{#1}}%
}

\author{%
  Duc Thien Nguyen \thanks{Co-first authors} \\
  FPT Software AI Center\\
  Hanoi, Vietnam \\
  \And
  Manh Duc Tuan Nguyen \printfnsymbol{1} \\
  FPT Software AI Center\\
  Hanoi, Vietnam \\
  \AND
  Truong Son Hy \printfnsymbol{1} \thanks{Correspondent author} \\
  Department of Mathematics and Computer Science \\
  Indiana State University \\
  Terre Haute, USA
  \And
  Risi Kondor \\
  Department of Computer Science \\
  University of Chicago \\
  Chicago, USA
}

\begin{document}

\maketitle




\begin{abstract}
Spatio-temporal signals forecasting plays an important role in numerous domains, especially in neuroscience and transportation. The task is challenging due to the highly intricate spatial structure, as well as the non-linear temporal dynamics of the network. To facilitate reliable and timely forecast for the human brain and traffic networks, we propose the \textit{Fast Temporal Wavelet Graph Neural Networks} (FTWGNN) that is both time- and memory-efficient for learning tasks on timeseries data with the underlying graph structure, thanks to the theories of \textit{multiresolution analysis} and \textit{wavelet theory} on discrete spaces. We employ \textit{Multiresolution Matrix Factorization} (MMF) \cite{pmlr-v32-kondor14} to factorize the highly dense graph structure and compute the corresponding sparse wavelet basis that allows us to construct fast wavelet convolution as the backbone of our novel architecture. Experimental results on real-world PEMS-BAY, METR-LA traffic datasets and AJILE12 ECoG dataset show that FTWGNN is competitive with the state-of-the-arts while maintaining a low computational footprint. Our PyTorch implementation is publicly available at \url{https://github.com/HySonLab/TWGNN}.
\end{abstract}
\section{Introduction} \label{sec:intro}

Time series modeling has been a quest in a wide range of academic fields and industrial applications, including neuroscience~\cite{POURAHMADI2016neuro} and traffic modeling~\cite{dcrnn}. Traditionally, model-based approaches such as autoregressive (AR) and Support Vector Regression~\cite{smola2004svr} require domain-knowledge as well as stationary assumption, which are often violated by the complex and non-linear structure of neural and traffic data. 

Recently, there has been intensive research with promising results on the traffic forecasting problem using deep learning such as Recurrent Neural Network (RNN)~\cite{qin2017rnn}, LSTM~\cite{koprinska2018lstm}, and graph-learning using Tranformer~\cite{xu2020spatial}. On the other hand, forcasting in neuroscience has been focusing mainly on long-term evolution of brain network structure based on fMRI data, such as predicting brain connectivities of an Alzheimer’s disease after several months~\cite{bessadok2022graph}, where existing methods are GCN-based~\cite{goktacs2020residual} or GAN-based graph autoencoder~\cite{gurler2020foreseeing}. Meanwhile, research on instantaneous time series forecasting of electroencephalogram (EEG) or 
electrocorticography (ECoG) remains untouched, even though EEG and ECoG are often cheaper and quicker to obtain than fMRI, while short-term forecasting may be beneficial for patients with strokes or epilepsy~\cite{shoeibi2022overview}.

In graph representation learning, a dense adjacency matrix expressing a densely connected graph can be a waste of computational resources, while physically, it may fail to capture the local ``smoothness'' of the network. To tackle such problems, a mathematical framework called Multiresolution Matrix Factorization (MMF) \cite{pmlr-v32-kondor14} has been adopted to ``sparsify'' the adjacency and graph Laplacian matrices of highly dense graphs. MMF is unusual amongst fast matrix factorization algorithms in that it does not make a low rank assumption. Multiresolution matrix factorization (MMF) is an alternative paradigm that is designed to capture structure at multiple different scales. This makes MMF especially well suited to modeling certain types of graphs with complex multiscale or hierarchical strucutre \cite{pmlr-v196-hy22a}, compressing hierarchical matrices (e.g., kernel/gram matrices) \cite{pmlr-v51-teneva16,NIPS2017_850af92f}, and other applications in computer vision \cite{8099564}. One important aspect of MMF is its ability to construct wavelets on graphs and matrices during the factorization process \cite{pmlr-v32-kondor14,pmlr-v196-hy22a}. The wavelet basis inferred by MMF tends to be highly sparse, that allows the corresponding wavelet transform to be executed efficiently via sparse matrix multiplication. \cite{pmlr-v196-hy22a} exploited this property to construct fast wavelet convolution and consequentially wavelet neural networks learning on graphs for graph classification and node classification tasks. In this work, we propose the incorporation of fast wavelet convolution based on MMF to build a time- and memory-efficient temporal architecture learning on timeseries data with the underlying graph structure.

From the aforementioned arguments, we propose the \textit{Fast Temporal Wavelet Graph Neural Network} (FTWGNN) for graph time series forecasting, in which the MMF theory is utilized to describe the local smoothness of the network as well as to accelerate the calculations. Experiments on real-world traffic and ECoG datasets show competitive performance along with remarkably smaller computational footprint of FTWGNN. In summary:
\begin{compactitem}
\item We model the spatial domain of the graph time series as a diffusion process, in which the theories of \textit{multiresolution analysis} and \textit{wavelet theory} are adopted. We employ \textit{Multiresolution Matrix Factorization} (MMF) to factorize the underlying graph structure and derive its sparse wavelet basis.
\item We propose the \textit{Fast Temporal Wavelet Graph Neural Network} (FTWGNN), an end-to-end model capable of modeling spatiotemporal structures.
\item We tested on two real-world traffic datasets and an ECoG dataset and achieved competitive results to state-of-the-art methods with remarkable reduction in computational time.
\end{compactitem}
\section{Related work} \label{sec:related}


A spatial-temporal forecasting task utilizes
spatial-temporal data information gathered from various sensors to predict their future states. Traditional approaches, such as the autoregressive integrated moving average (ARIMA), k-nearest neighbors algorithm (kNN), and support vector machine (SVM), can only take into account temporal information without considering spatial
features~\cite{van2012short,jeong2013supervised}. Aside from traditional approaches, deep neural networks are proposed to model much more complex spatial-temporal relationships. Specifically, by using an extended fully-connected LSTM with embedded convolutional layers, FC-LSTM \cite{sutskever2014sequence} specifically combines CNN and LSTM to model spatial and temporal relations. When predicting traffic, ST-ResNet~\cite{zhang2017deep} uses a deep residual CNN network, revealing the powerful capabilities of the residual network. Despite the impressive results obtained, traffic forecasting scenarios with graph-structured data is incompatible with all of the aforementioned methods because they are built for grid data. For learning tasks on graphs, node representations in GNNs~\cite{Kipf_GCN} uses a neighborhood aggregation scheme, which involves sampling and aggregating the features of nearby nodes. Since temporal-spatial data such as traffic data or brain network is a well-known type of non-Euclidean structured graph data, great efforts have been made to use graph convolution methods in traffic forecasting. As an illustration, DCRNN \cite{dcrnn} models traffic flow as a diffusion process and uses directed graph bidirectional random walks to model spatial dependency.

In the field of image and signal processing, processing is more efficient and simpler in a sparse representation where fewer coefficients reveal the information that we are searching for. Based on this motivation, Multiresolution Analysis (MRA) has been proposed by \cite{192463} as a design for multiscale signal approximation in which the sparse representations can be constructed by decomposing signals over elementary waveforms chosen in a family called \textit{wavelets}. Besides Fourier transforms, the discovery of wavelet orthogonal bases such as Haar \cite{Haar1910ZurTD} and Daubechies \cite{Daubechies1988OrthonormalBO} has opened the door to new transforms such as continuous and discrete wavelet transforms and the fast wavelet transform algorithm that have become crucial for several computer applications \cite{10.5555/1525499}.

\cite{pmlr-v32-kondor14} and \cite{pmlr-v196-hy22a} have introduced Multiresolution Matrix Factorization (MMF) as a novel method for constructing sparse wavelet transforms of functions defined on the nodes of an arbitrary graph while giving a multiresolution approximation of hierarchical matrices. MMF is closely related to other works on constructing wavelet bases on discrete spaces, including wavelets defined based on diagonalizing the diffusion operator or the normalized graph Laplacian \cite{COIFMAN200653} \cite{HAMMOND2011129} and multiresolution
on trees \cite{10.5555/3104322.3104370} \cite{10.2307/30244209}.
\section{Background} \label{sec:background}

\subsection{Multiresolution Matrix Factorization}

Most commonly used matrix factorization algorithms, such as principal component analysis (PCA), singular value decomposition (SVD), or non-negative matrix factorization (NMF) are inherently single-level algorithms. A symmetric matrix $\mA \in \mathbb{R}^{n \times n}$ is of rank $r \ll n$ if it can be expressed in terms of a dictionary of $r$ mutually orthogonal unit vectors $\{u_1, u_2, \dots, u_r\}$ in the form
$$\mA = \sum_{i = 1}^r \lambda_i u_i u_i^T,$$
where $u_1, \dots, u_r$ are the normalized eigenvectors of $A$ and $\lambda_1, \dots, \lambda_r$ are the corresponding eigenvalues. This is the decomposition that PCA finds, and it corresponds to factorizing $\mA$ in the form
\begin{equation}
\mA = \mU^T \mH \mU,
\label{eq:eigen}
\end{equation}
where $\mU$ is an orthogonal matrix and $\mH$ is a diagonal matrix with the eigenvalues of $\mA$ on its diagonal. The drawback of PCA is that eigenvectors are almost always dense, while matrices occuring in learning problems, especially those related to graphs, often have strong locality properties, in the sense that they are more closely couple certain clusters of nearby coordinates than those farther apart with respect to the underlying topology. In such cases, modeling $A$ in terms of a basis of global eigenfunctions is both computationally wasteful and conceptually unreasonable: a localized dictionary would be more appropriate. In contrast to PCA, \cite{pmlr-v32-kondor14} proposed \textit{Multiresolution Matrix Factorization}, or MMF for short, to construct a sparse hierarchical system of $L$-level dictionaries. The corresponding matrix factorization is of the form
$$\mA = \mU_1^T \mU_2^T \dots \mU_L^T \mH \mU_L \dots \mU_2 \mU_1,$$
where $\mH$ is close to diagonal and $\mU_1, \dots, \mU_L$ are sparse orthogonal matrices such that:
\begin{compactitem}
\item Each $\mU_\ell$ is $k$-point rotation (i.e. Givens rotation) for some small $k$, meaning that it only rotates $k$ coordinates at a time. Formally, Def.~\ref{def:rotation-matrix} defines the $k$-point rotation matrix. 
\item There is a nested sequence of sets $\sS_L \subseteq \cdots \subseteq \sS_1 \subseteq \sS_0 = [n]$ such that the coordinates rotated by $\mU_\ell$ are a subset of $\sS_\ell$.
\item $\mH$ is an $\sS_L$-core-diagonal matrix that is formally defined in Def.~\ref{def:core-diagonal}.
\end{compactitem}

We formally define MMF in Defs.~\ref{def:mmf} and \ref{def:factorizable}. A special case of MMF is the Jacobi eigenvalue algorithm \cite{Jacobi+1846+51+94} in which each $\mU_\ell$ is a 2-point rotation (i.e. $k = 2$).

\subsection{Multiresolution analysis} \label{sec:multiresolution-analysis}

\cite{pmlr-v32-kondor14} has shown that MMF mirrors the classical theory of multiresolution analysis (MRA) on the real line \cite{192463} to discrete spaces. The functional analytic view of wavelets is provided by MRA, which, similarly to Fourier analysis, is a way of filtering some function space into a sequence of subspaces
\begin{equation}
\dots \subset \sV_{-1} \subset \sV_0 \subset \sV_1 \subset \sV_2 \subset \dots
\label{eq:subspace-sequence}
\end{equation}
However, it is best to conceptualize (\ref{eq:subspace-sequence}) as an iterative process of splitting each $\sV_\ell$ into the orthogonal sum $\sV_\ell = \sV_{\ell + 1} \oplus \sW_{\ell + 1}$ of a smoother part $\sV_{\ell + 1}$, called the \textit{approximation space}; and a rougher part $\sW_{\ell + 1}$, called the \textit{detail space} (see Fig.~\ref{fig:subspaces}). Each $\sV_\ell$ has an orthonormal basis $\Phi_\ell \triangleq \{\phi_m^\ell\}_m$ in which each $\phi$ is called a \textit{father} wavelet. Each complementary space $\sW_\ell$ is also spanned by an orthonormal basis $\Psi_\ell \triangleq \{\psi_m^\ell\}_m$ in which each $\psi$ is called a \textit{mother} wavelet. In MMF, each individual rotation $\mU_\ell: \sV_{\ell - 1} \rightarrow \sV_\ell \oplus \sW_\ell$ is a sparse basis transform that expresses $\Phi_\ell \cup \Psi_\ell$ in the previous basis $\Phi_{\ell - 1}$ such that:
$$\phi_m^\ell = \sum_{i = 1}^{\text{dim}(\sV_{\ell - 1})} [\mU_\ell]_{m, i} \phi_i^{\ell - 1}, \ \ \ \ \psi_m^\ell = \sum_{i = 1}^{\text{dim}(\sV_{\ell - 1})} [\mU_\ell]_{m + \text{dim}(\sV_{\ell - 1}), i} \phi_i^{\ell - 1},$$
in which $\Phi_0$ is the standard basis, i.e. $\phi_m^0 = e_m$; and $\text{dim}(\sV_\ell) = d_\ell = \lvert \sS_\ell \rvert$. In the $\Phi_1 \cup \Psi_1$ basis, $\mA$ compresses into $\mA_1 = \mU_1\mA\mU_1^T$. In the $\Phi_2 \cup \Psi_2 \cup \Psi_1$ basis, it becomes $\mA_2 = \mU_2\mU_1\mA\mU_1^T\mU_2^T$, and so on. Finally, in the $\Phi_L \cup \Psi_L \cup \dots \cup \Psi_1$ basis, it takes on the form $\mA_L = \mH = \mU_L \dots \mU_2\mU_1 \mA \mU_1^T\mU_2^T \dots \mU_L^T$ that consists of four distinct blocks (supposingly that we permute the rows/columns accordingly):
$$\mH = \begin{pmatrix} \mH_{\Phi, \Phi} & \mH_{\Phi, \Psi} \\ \mH_{\Psi, \Phi} & \mH_{\Psi, \Psi} \end{pmatrix},$$
where $\mH_{\Phi, \Phi} \in \mathbb{R}^{\text{dim}(\sV_L) \times \text{dim}(\sV_L)}$ is effectively $\mA$ compressed to $\sV_L$, $\mH_{\Phi, \Psi} = \mH_{\Psi, \Phi}^T = 0$ and $\mH_{\Psi, \Psi}$ is diagonal. MMF approximates $\mA$ in the form
$$\mA \approx \sum_{i, j = 1}^{d_L} h_{i, j} \phi_i^L {\phi_j^L}^T + \sum_{\ell = 1}^L \sum_{m = 1}^{d_\ell} c_m^\ell \psi_m^\ell {\psi_m^\ell}^T,$$
where $h_{i, j}$ coefficients are the entries of the $\mH_{\Phi, \Phi}$ block, and $c_m^\ell = \langle \psi_m^\ell, \mA \psi_m^\ell \rangle$ wavelet frequencies are the diagonal elements of the $\mH_{\Psi, \Psi}$ block.

In particular, the dictionary vectors corresponding to certain rows of $\mU_1$ are interpreted as level one wavelets, the dictionary vectors corresponding to certain rows of $\mU_2\mU_1$ are interpreted as level two wavelets, and so on. One thing that is immediately clear is that whereas Eq.~(\ref{eq:eigen}) diagonalizes $\mA$ in a single step, multiresolution analysis will involve a sequence of basis transforms $\mU_1, \mU_2, \dots, \mU_L$, transforming $\mA$ step by step as
\begin{equation}
\mA \rightarrow \mU_1\mA\mU_1^T \rightarrow \dots \rightarrow \mU_L \dots \mU_1\mA\mU_1^T \dots \mU_L^T \triangleq \mH,
\label{eq:mmf-transform}
\end{equation}
so the corresponding matrix factorization must be a multilevel factorization
\begin{equation}
\mA \approx \mU_1^T \mU_2^T \dots \mU_\ell^T \mH \mU_\ell \dots \mU_2 \mU_1.
\label{eq:mmf-factorization}
\end{equation}
\section{Method} \label{sec:method}
\begin{figure*}
  \centering
    \includegraphics[width=0.9\textwidth]{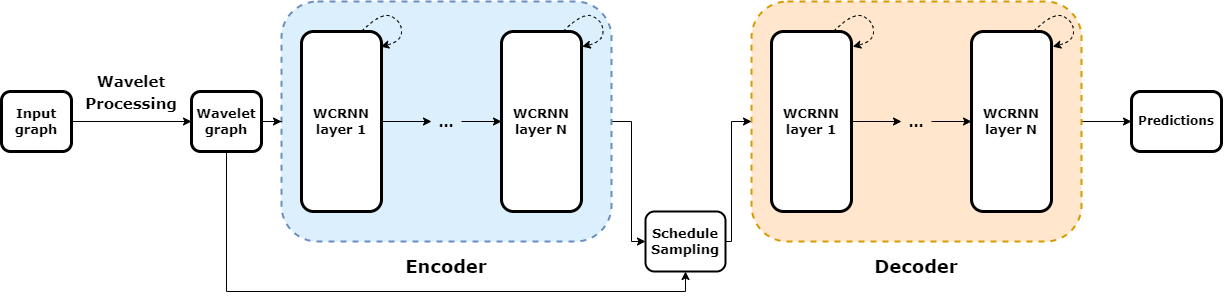}
    \centering
    \caption{Architecture of Fast Temporal Wavelet Neural Network. \textbf{WC:} graph wavelet convolution given MMF's wavelet basis.}
  \label{fig:architecture}    
\end{figure*}

\subsection{Wavelet basis and convolution on graph}

Section \ref{sec:multiresolution-analysis} introduces the theory of multiresolution analysis behind MMF as well as the construction of a \textit{sparse} wavelet basis for a symmetric matrix $\mA \in \mathbb{R}^{n \times n}$. Without the loss of generality, we assume that $\mA$ is a weight matrix of a weighted undirected graph $\mathcal{G} = (V, E)$ in which $V = \{v_1, .., v_n\}$ is the set of vertices and $E = \{(v_i, v_j)\}$ is the set of edges with the weight of edge $(v_i, v_j)$ is given by $\mA_{i, j}$. Given a graph signal $\vf \in \mathbb{R}^n$ that is understood as a function $f: V \rightarrow \mathbb{R}$ defined on the vertices of the graph, the wavelet transform (up to level $L$) expresses this graph signal, without loss of generality $f \in \sV_0$, as:
$$f(v) = \sum_{\ell = 1}^L \sum_m \alpha_m^\ell \psi_m^\ell(v) + \sum_m \beta_m \phi_m^L(v), \ \ \ \ \text{for each} \ \ v \in V,$$ 
where $\alpha_m^\ell = \langle f, \psi_m^\ell \rangle$ and $\beta_m = \langle f, \phi_m^L \rangle$ are the wavelet coefficients. Based on the wavelet basis construction via MMF detailed in \cite{pmlr-v196-hy22a}:
\begin{compactitem}
\item For $L$ levels of resolution, we get exactly $L$ mother wavelets $\overline{\psi} = \{\psi^1, \psi^2, \dots, \psi^L\}$, each corresponds to a resolution.
\item The rows of $\mH = \mA_L$ make exactly $n - L$ father wavelets $\overline{\phi} = \{\phi^L_m = \mH_{m, :}\}_{m \in \sS_L}$. In total, a graph of $n$ vertices has exactly $n$ wavelets, both mothers and fathers.
\end{compactitem}
Analogous to the convolution based on Graph Fourier Transform \cite{ae482107de73461787258f805cf8f4ed}, each convolution layer $k \in \{1, .., K\}$ of wavelet neural network transforms an input vector $\vf^{(k - 1)}$ of size $\lvert V \rvert \times F_{k - 1}$ into an output $\vf^{(k)}$ of size $\lvert V \rvert \times F_k$ as
\begin{equation}
\vf^{(k)}_{:, j} = \sigma \bigg( \mW \sum_{i = 1}^{F_{k - 1}} \vg^{(k)}_{i, j} \mW^T \vf^{(k - 1)}_{:, i} \bigg) \ \ \ \ \text{for} \ \ j = 1, \dots, F_k,
\label{eq:wavevlet-conv}
\end{equation}
where $\mW$ is our wavelet basis matrix as we concatenate $\overline{\phi}$ and $\overline{\psi}$ column-by-column, $\vg^{(k)}_{i, j}$ is a parameter/filter in the form of a diagonal matrix learned in spectral domain, and $\sigma$ is an element-wise non-linearity (e.g., ReLU, sigmoid, etc.). In Eq.(\ref{eq:wavevlet-conv}), first we employ the wavelet transform of a graph signal $\vf$ into the spectral domain (i.e. $\hat{\vf} = \mW^T \vf$ is the forward transform and $\vf = \mW \hat{\vf}$ is the inverse transform), then a learnable filter $\vg$ to the wavelet coefficients, the inverse transform back to the spatial domain and let everything through a non-linearity $\sigma$. Since the wavelet basis matrix $\mW$ is \textit{sparse}, both the wavelet transform and its inverse transform can be implemented efficiently via sparse matrix multiplication. 

\subsection{Temporal Wavelet Neural Networks}


Capturing spatiotemporal dependencies among time series in various spatiotemporal forecasting problems demands both spatial and temporal models. We build our novel \textit{Fast Temporal Wavelet Graph Neural Network} with the architectural backbone from \textit{Diffusion Convolutional Recurrent Neural Network} (DCRNN) \cite{dcrnn}, that combines both spatial and temporal models to solve these tasks.

\textbf{Spatial Dependency Model} \quad The spatial dynamic in the network is captured by diffusion process. Let $G = (\mX,\mA)$ represent an undirected graph, where $\mX =[\vx_{1}^T,\dots,\vx_{N}^T]^T \in \mathbb{R}^{N \times D}$ denotes signals of $N$ nodes, each has $D$ features. Define further the right-stochastic edge weights matrix $\Tilde{\mA} \in \mathbb{R}^{N \times N}$  in which $\sum_j \Tilde{\mA}_{ij}=1 \forall i$. In the simplest case, when $\Tilde{\mL}=\mI-\Tilde{\mA}$ is the nomalized random walk matrix, the diffusion process on graph is governed by the following equation \cite{GRAND}:
\begin{equation}
    \frac{{\rm d} \mX(t)}{{\rm d}t} = (\Tilde{\mA}-\mI)\mX(t)
    \label{eq:diffusion}
\end{equation}
where $\mX(t) = [\vx_{1}^T(t),\dots,[\vx_{N}^T(t)]^T \in \mathbb{R}^{N \times D}$ and $\mX(0)=\mX$. Applying forward Euler discretization with step size 1, gives:

\begin{equation}
\mX(k) = \mX(k-1) + (\Tilde{\mA}-\mI)\mX(k-1) = \mX(k-1) -\Tilde{\mL}\mX(k-1) = \Tilde{\mA}\mX(k-1) = \Tilde{\mA}^k\mX(0)
\label{eq:random_walk}
\end{equation}

Eq.\ref{eq:random_walk} is similar to the well-established GCN architecture propose in \cite{Kipf_GCN}. Then, the diffusion convolution operation over a graph signal $\mathbb{R}^{N \times D}$ and filter $f_{\boldsymbol{\theta}}$ is defined as:
\begin{equation}
    \mX_{:,d}\star_\mathcal{G}f_{\boldsymbol{\theta}} = \sum_{k=0}^{K-1}\theta_k \Tilde{\mA}^k \mX_{:,d} \quad \forall d \in \{1,\dots,D\}
    \label{eq:graph_conv}
\end{equation}
where $\mTheta \in \mathbb{R}^{K \times 2}$ are the parameters for the filter.

\textbf{Temporal Dependency Model} \quad The DCRNN is leveraged from the recurrent neural networks (RNNs) to model the temporal dependency. In particular, the matrix multiplications in GRU is replaced with the diffusion convolution, which is called \textit{Diffusion Convolutional Gated Recurrent Unit} (DCGRU).
\begin{align}
    \vr^{(t)} &= \sigma(\mTheta_r \star_\mathcal{G} [\mX^{(t)}, \mH^{(t-1)}] + \boldsymbol{b} ) \notag \\
    \boldsymbol{u}^{(t)} &= \sigma(\mTheta_r \star_\mathcal{G} [\mX^{(t)}, \mH^{(t-1)}] + \boldsymbol{b_u} ) \notag \\
    \mC^{(t)} &= \tanh(\mTheta_r \star_\mathcal{G} [\mX^{(t)}, (\boldsymbol{r} \odot \mH^{(t-1)})] + \boldsymbol{b_c} ) \notag \\
    \mH^{(t)} &= \boldsymbol{u}^{(t)} \odot \mH^{(t-1)} + (1-\boldsymbol{u}^{(t)}) \odot \mC^{(t)} \notag
\end{align}
where $\mX(t), \mH(t)$ denote the input and output of at time $t$, while $\vr^{(t)},\boldsymbol{u}^{(t)}$ are reset gate and update gate at time $t$, respectively. 

Both the encoder and the decoder are recurrent neural networks with DCGRU following \textit{Sequence-to-Sequence} style. To mitigate the distribution differences between training and testing data, scheduled sampling technique \cite{bengio2015scheduled} is used, where the model is fed with either the ground truth with probability $\epsilon_i$ or the prediction by the model with probability $1-\epsilon_i$. 


For our novel \textit{Fast Temporal Wavelet Graph Neural Network} (FTWGNN), the fundamental difference is that instead of using temporal traffic graph as the input of DCRNN, we use the sparse wavelet basis matrix $\mW$ which is extracted via MMF (see Section \ref{sec:multiresolution-analysis}) and replace the diffusion convolution by our fast \textit{wavelet convolution}. Given the sparsity of our wavelet basis, we significantly reduce the overall computational time and memory usage. Each Givens rotation matrix $\mU_\ell$ (see Def.~\ref{def:rotation-matrix}) is a highly-sparse orthogonal matrix with a non-zero core of size $K \times K$. The number of non-zeros in MMF's wavelet basis $\mW$, that can be computed as product $\mU_1\mU_2 \cdots \mU_L$, is $O(LK^2)$ where $L$ is the number of resolutions (i.e. number of Givens rotation matrices) and $K$ is the number of columns in a Givens rotation matrix.  \cite{pmlr-v32-kondor14} and \cite{pmlr-v196-hy22a} have shown in both theory and practice that $L$ only needs to be in $O(n)$ where $n$ is the number of columns and $K$ small (e.g., 2, 4, 8) to get a decent approximation/compression for a symmetric hierarchical matrix. Technically, MMF is able to compress a symmetric hierararchical matrix from the original quadratic size $n \times n$ to a linear number of non-zero elements $O(n)$. Practically, all the Givens rotation matrices $\{\mU_\ell\}_{\ell = 1}^L$ and the wavelet basis $\mW$ can be stored in Coordinate Format (COO), and the wavelet transform and its inverse in wavelet convolution (see Eq.~\ref{eq:wavevlet-conv}) can be implemented efficiently by sparse matrix multiplication in PyTorch's sparse library \cite{paszke2019pytorch}. The architecture of our model is shown in Figures \ref{fig:architecture} and \ref{fig:architecture2}.

\section{Experiments} \label{sec:experiments}

Our PyTorch implementation is publicly available at \url{https://github.com/HySonLab/TWGNN}. The implementation of multiresolution matrix factorization and graph wavelet computation \cite{pmlr-v196-hy22a} is publicly available at \url{https://github.com/risilab/Learnable_MMF}.

To showcase the competitive performance and remarkable acceleration of FTWGNN, we conducted experiments on two well-known traffic forecasting benchmarks METR-LA and PEMS-BAY, and one challenging ECoG dataset AJILE12. We compare our model with widely used time series models, including:
\begin{enumerate*}
    \item HA: Historical Average;
    \item $\text{ARIMA}_{kal}$;
    \item VAR~\cite{hamilton2020time};
    \item SVR~\cite{smola2004svr}; 
    \item FNN; 
    \item FC-LSTM~\cite{sutskever2014sequence};
    \item STGCN~\cite{han2020stgcn};
    \item GWaveNet~\cite{wu2019graph}; and
    \item DCRNN~\cite{dcrnn}.
\end{enumerate*}
Methods are evaluated on three metrics:
\begin{enumerate*}[label= \textbf{(\roman*)}]
    \item Mean Absolute Error (MAE);
    \item Mean Absolute Percentage Error (MAPE); and
    \item Root Mean Squared Error (RMSE).
\end{enumerate*}
FTWGNN and DCRNN are implemented using PyTorch~\cite{paszke2019pytorch} on an NVIDIA A100-SXM4-80GB GPU. Detailed settings of FTWGNN and data preparation are in \ref{app:ftwgnn}.

\begin{table*}[hbt!]
\centering
    \resizebox{\textwidth}{!}{\begin{tabular}{c||c|c|cccccccccc}
         \hline
          Dataset & $T$ & Metric & HA & $\text{ARIMA}_{kal}$ & VAR & SVR & FNN & FC-LSTM & STGCN & GWaveNet & DCRNN & FTWGNN \\  
         \hline 
         \multirow{10}{*}{METR-LA}&&MAE & 4.16 & 3.99 &4.42 & 3.99 & 3.99 & 3.44 & 2.88 & \textbf{2.69} & 2.77 & {2.70}\\
         &15 min &RMSE& 7.80 & 8.21 & 7.89 & 8.45 & 7.94 & 6.30 & 5.74 & \textbf{5.15} & 5.38 & \textbf{5.15}\\
         &&MAPE& 13.0\% & 9.6\% & 10.2\% & 9.3\% & 9.9\% & 9.6\% & 7.6\% & 6.9\% & 7.3\% & \textbf{6.8\%}\\
         \cline{2-13} 
         &&MAE& 4.16 & 5.15 & 5.41 & 5.05 & 4.23 & 3.77 & 3.47 & 3.07 & 3.15 & \textbf{3.02}\\
         &30 min &RMSE& 7.80 & 10.45 & 9.13 & 10.87 & 8.17 & 7.23 & 7.24 & 6.22 & 6.45 & \textbf{5.95}\\
         &&MAPE& 13.0\% & 12.7\% & 12.7\% & 12.1\% & 12.9\% & 10.9\% & 9.6\% & 8.4\% & 8.8\% & \textbf{8.0\%}\\
        \cline{2-13} 
         &&MAE& 4.16 & 6.90 & 6.52 & 6.72 & 4.49 & 4.37 & 4.59 & 3.53 & 3.60 & \textbf{3.42}\\
         &60 min &RMSE& 7.80 & 13.23 & 10.11 & 13.76 & 8.69 & 8.69 & 9.40 & 7.37 & 7.59 & \textbf{6.92}\\
         &&MAPE& 13.0\% & 17.4\% & 15.8\% & 16.7\% & 14.0\% & 13.2\% & 12.7\% & 10.0\% & 10.5\% &\textbf{9.8\%}\\
        \hline \hline
         \multirow{10}{*}{PEMS-BAY}&&MAE& 2.88 & 1.62 & 1.74 & 1.85 & 2.20 & 2.05 & 1.36 & 1.3 & 1.38 & \textbf{1.14}\\
         &15 min &RMSE& 5.59 & 3.30 & 3.16 & 3.59 & 4.42 & 4.19 & 2.96 & 2.74 & 2.95 & \textbf{2.40}\\
         &&MAPE& 6.8\% & 3.5\% & 3.6\% & 3.8\% & 5.2\% & 4.8\% & 2.9\% & 2.7\% & 2.9\% & \textbf{2.3\%}\\
         \cline{2-13} 
         &&MAE& 2.88 & 2.33 & 2.32 & 2.48 & 2.30 & 2.20 & 1.81 & 1.63 & 1.74 & \textbf{1.50}\\
         &30 min &RMSE& 5.59 & 4.76 & 4.25 & 5.18 & 4.63 & 4.55 & 4.27 & 3.70 & 3.97 & \textbf{3.27}\\
         &&MAPE& 6.8\% & 5.4\% & 5.0\% & 5.5\% & 5.43\% & 5.2\% & 4.2\% & 3.7\% & 3.9\% & \textbf{3.2\%}\\
        \cline{2-13} 
         &&MAE& 2.88 & 3.38 & 2.93 & 3.28 & 2.46 & 2.37 & 2.49 & 1.95 & 2.07 & \textbf{1.79}\\
         &60 min &RMSE& 5.59 & 6.5 & 5.44 & 7.08 & 4.98 & 4.96 & 5.69 & 4.52 & 4.74 & \textbf{3.99}\\
         &&MAPE& 6.8\% & 8.3\% & 6.5\% & 8.0\% & 5.89\% & 5.7\% & 5.8\% & 4.6\% & 4.9\% & \textbf{4.1\%}\\
        \hline
    \end{tabular}}
    \caption{Performance comparison of different models for traffic speed forecasting.}
      \vspace{-0.3cm}
    \label{tab:result_acc}
\end{table*}

\begin{table}
\centering
    \small
    \begin{tabular}{c||c|cc|c}
         \hline
          Dataset & $T$ & DCRNN & FTWGNN & Speedup \\  
         \hline 
         \multirow{3}{*}{METR-LA}&15 min& 350s & \textbf{217s} & 1.61x \\
         \cline{2-5}
         & 30 min& 620s & \textbf{163s} & 3.80x \\
         \cline{2-5}
         & 60 min& 1800s & \textbf{136s} & 13.23x \\
         \hline
         \multirow{3}{*}{PEMS-BAY}&15 min& 427s & \textbf{150s} & 2.84x \\
         \cline{2-5}
         & 30 min& 900s & \textbf{173s} & 5.20x \\
         \cline{2-5}
         & 60 min& 1800s & \textbf{304s} & 5.92x \\
         \hline
         \multirow{3}{*}{AJILE12}&1 sec& 80s & \textbf{35s} & 2.28x \\
         \cline{2-5}
         &5 sec& 180s & \textbf{80s} & 2.25x \\
         \cline{2-5}
         & 15 sec& 350s & \textbf{160s} & 2.18x \\
         \hline

    \end{tabular}
    \caption{Training time/epoch between DCRNN and FTWGNN.}
    \label{tab:trainTime}
    \vspace{-0.5cm}
\end{table}



\textbf{Adjacency matrix} \quad According to DCRNN~\cite{dcrnn}, the traffic sensor network is expressed by an adjacency matrix which is constructed using the Gaussian kernel thresholded~\cite{shuman2013signal}. Specifically, for each pair of sensors $v_i$ and $v_j$, the edge weight of $v_i$ to $v_j$, denoted by $A_{ij}$, is defined as
\begin{equation} \label{eq:adjacency}
    A_{ij} \coloneqq
    \left
    \{
        \begin{aligned}
        & \exp( -\frac{\text{dist}(v_i, v_j)}{\sigma^2} ), \quad & \text{dist}(v_i, v_j) \leq k \\
        & 0, \quad & \text{otherwise}
        \end{aligned}
    \right. \,,
\end{equation}
where $\text{dist}(v_i, v_j)$ denotes the spatial distance from $v_i$ to $v_j$, $\sigma$ is the standard deviation of the distances and $k$ is the distance threshold. 

Nevertheless, such a user-defined adjacency matrix requires expert knowledge, and thus may not work on other domains, \eg, brain networks. In the ECoG time series forecasting case, the adjacency matrix is computed based on the popular Local Linear Embedding (LLE)~\cite{saul2003think}. In particular, for the matrix data $\mX=[\vx_1, \ldots, \vx_N] \in \sR^{T \times N}$ where $\vx_i$ denotes the time series data of node $i$ for $i \in \{1, \ldots, N\}$, an adjacency matrix $\mA$ is identified to gather all the coefficients of the affine dependencies among $\{\vx_i\}_{i=1}^N$ by solving the following optimization problem.
\begin{align} 
  \mA \coloneqq \arg\min_{\hat{\mA} \in \sR^{N \times N}}
  {} & {} \norm{ \mX - \mX \hat{\mA}^T}_{\text{F}}^2 + \lambda_{A} \norm{ \hat{\mA} }_1 \notag \\
  \text{s.to} \quad 
  {} & {} \vone^T_N \hat{\mA} = \vone^T_N\,,
       \quad \text{diag}(\hat{\mA}) = \vzero \,, \label{manifold.identify}
\end{align}
where the constraint $\vone^T_N \hat{\mA} = \vone^T_N$ realizes the affine combinations, while 
$\text{diag}(\hat{\mA}) = \vzero$ excludes the trivial solution $\hat{\mA} = \mI_N$. Furthermore, to promote the local smoothness of the graph, each data point $x_i$ is assumed to be approximated by a few neighbors $\{x_{j_1}, x_{j_2}, \ldots, x_{j_k}\}$, thus $\hat{A}$ is regularized by the $l_1$-norm loss $\norm{ \hat{\mA} }_1$ to be sparse.
Task (\ref{manifold.identify}) is a composite convex minimization problem with affine constraints, which can therefore be solved by~\cite{slavakis2018fejer}.

\textbf{Construction of wavelet bases} \quad For traffic data sets, 100 mother wavelets are extracted, \ie, $L=100$, while for the AJILE12 dataset, $L=10$. The sparsity of wavelet bases is reported in Table \ref{tab:basisDensity}, which demonstrate a remarkable compression of wavelet bases compared to that of Fourier bases.
\begin{table}[!ht]
\centering
    \begin{tabular}{|c|cc|}
         \hline
         \textbf{Dataset} & \textbf{Fourier basis} & \textbf{Wavelet basis} \\  
         \hline
         \hline
         METR-LA & 99.04\% & \textbf{1.11\%} \\
         \hline
         PEMS-BAY & 96.35\% & \textbf{0.63\%} \\
         \hline
         AJILE12 & 100\% & \textbf{1.81\%} \\
         \hline

    \end{tabular}
    \caption{Sparsity bases (i.e. percentage of non-zeros).}
    \label{tab:basisDensity}
    \vspace{-0.5cm}
\end{table}

\subsection{Traffic prediction} \label{sec:traffic}

Two real-world large-scale traffic datasets are considered:
\begin{compactitem}
    \item \textbf{METR-LA} Data of 207 sensors in the highway of Los Angeles County~\cite{metrladata} over the 4-month period from Mar 1st 2012 to Jun 30th 2012.
    \item \textbf{PEMS-BAY} Data of 325 sensors in the Bay Area over the 6-month period from Jan 1st 2017 to May 31th 2017 from the California Transportation Agencies (CalTrans) Performance Measurement System (PeMS).
\end{compactitem}

    

\begin{figure*}[!ht]
\begin{subfigure}{.5\textwidth}
    \centering
    \includegraphics[width=1.0\linewidth]{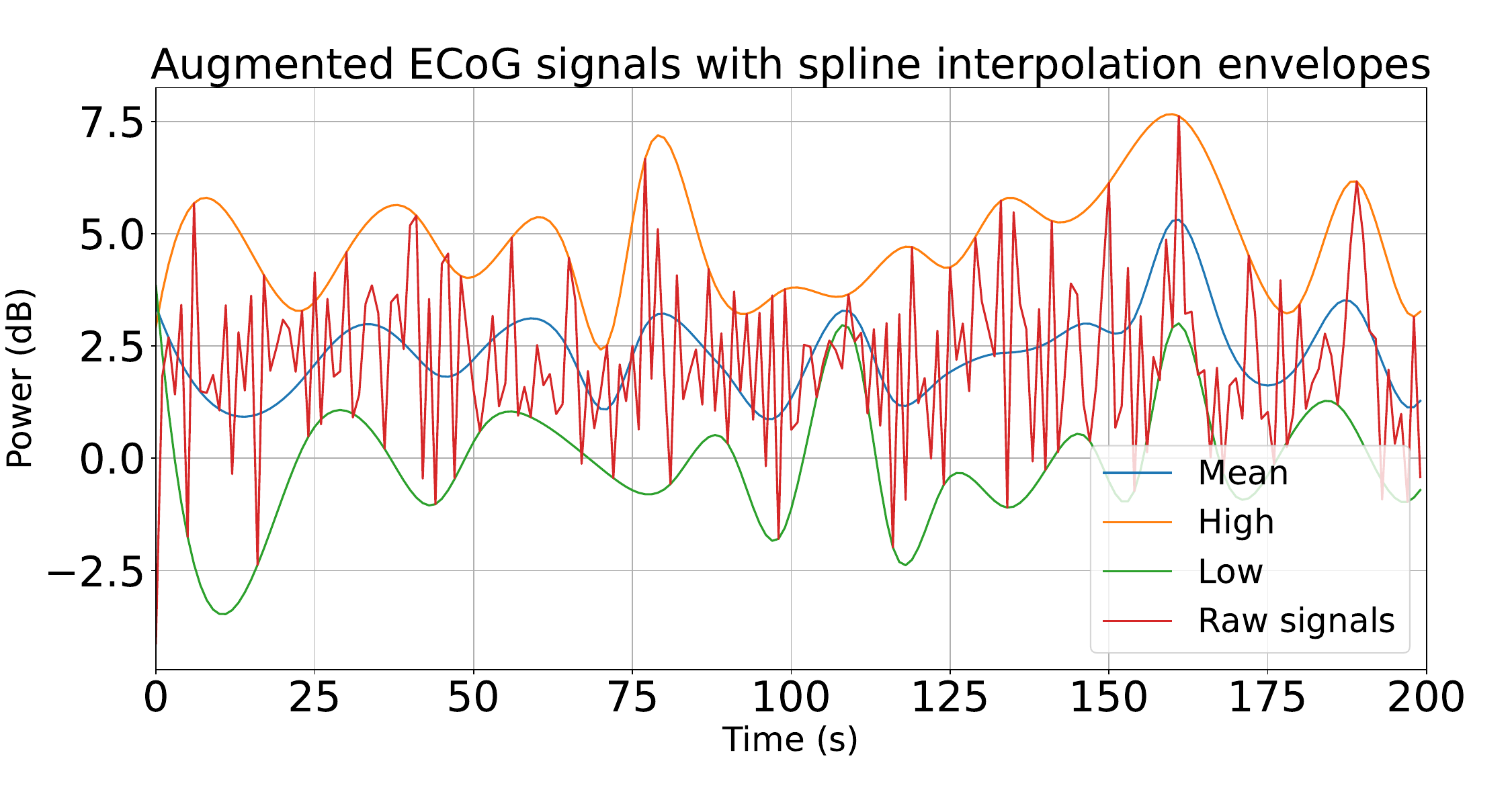}
    {\phantomcaption\ignorespaces\label{fig:augmentedECG}}
    \vspace{-5pt}
\end{subfigure}%
\begin{subfigure}{.5\textwidth}
    \centering
    \includegraphics[width=1.0\linewidth]{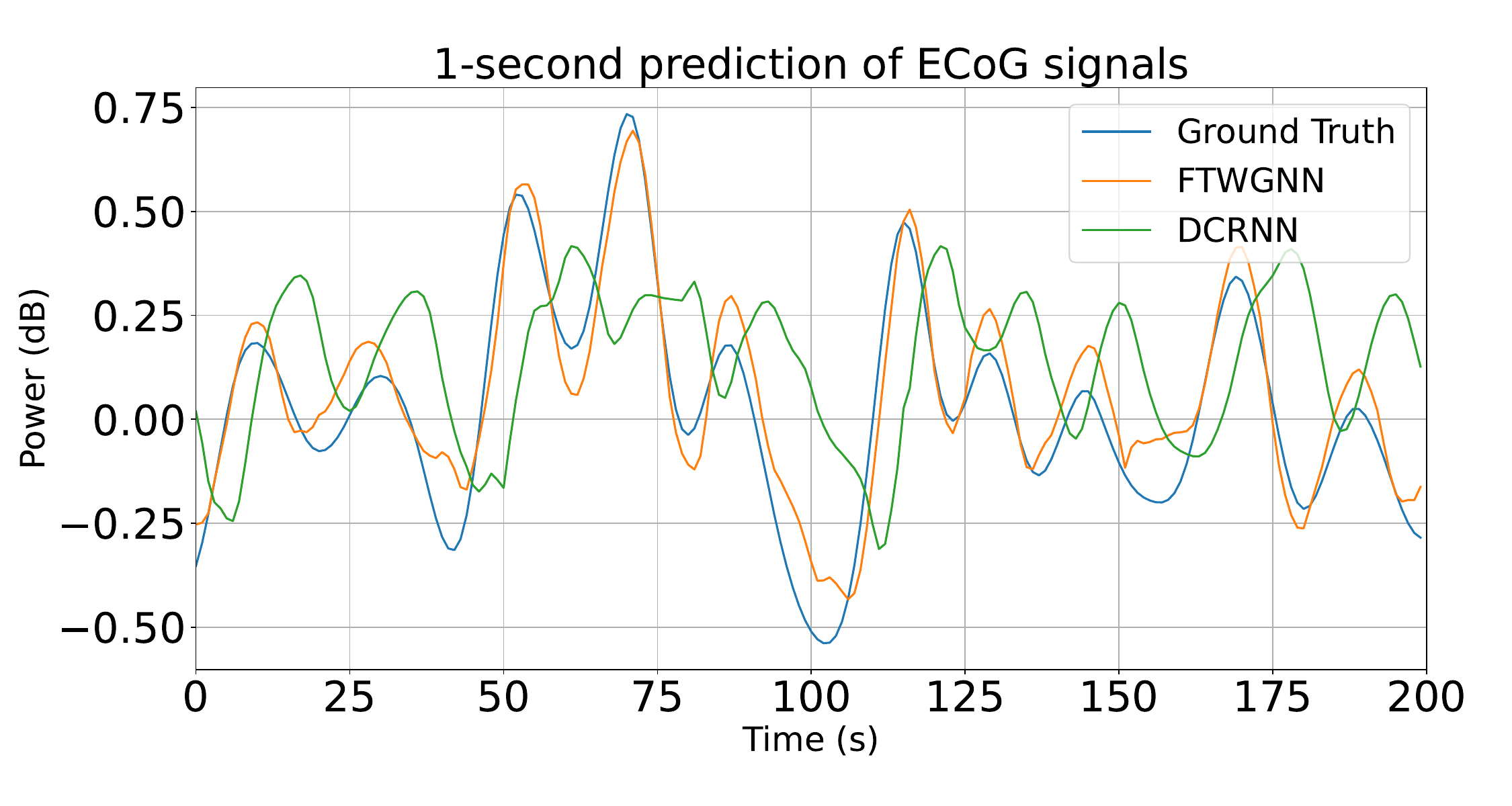}
    {\phantomcaption\ignorespaces\label{fig:expECG1s}}
    \vspace{-5pt}
\end{subfigure}
\caption{(a) Augmented ECoG signals by spline interpolation envelopes, (b) 1-second prediction of ECoG signals}
\vspace{-5pt}
\end{figure*}

\begin{table*}[!hbt]
\centering
    \resizebox{0.7\textwidth}{!}{\begin{tabular}{c||c|c|ccccccc}
         \hline
          Dataset & $T$ & Metric & HA & VAR & LR & SVR & LSTM & DCRNN & FTWGNN \\  
         \hline 
         \multirow{10}{*}{AJILE12}&&MAE& 0.88 & 0.16 & 0.27 & 0.27 & 0.07 & 0.05 & \textbf{0.03} \\
         &1 sec &RMSE& 1.23 & 0.25 & 0.37 & 0.41 & \textbf{0.09} & 0.45 & 0.35 \\
         &&MAPE& 320\% & 58\% & 136\% & 140\% & 38\% & 7.84\% & \textbf{5.27\%} \\
         \cline{2-10} 
         &&MAE & 0.88 & 0.66 & 0.69 & 0.69 & 0.39 & 0.16 & \textbf{0.11}\\
         &5 sec &RMSE& 1.23 & 0.96 & 0.92 & 0.93 & 0.52 & 0.24 & \textbf{0.15} \\
         &&MAPE& 320\% & 221\% & 376\% & 339\% & 147\% & 64\% & \textbf{57\%}\\
         \cline{2-10} 
         &&MAE& 0.88 & 0.82 & 0.86 & 0.86 & 0.87 & 0.78 & \textbf{0.70} \\
         &15 sec &RMSE& 1.23 & 1.15 & 1.13 & 1.13 & 1.14 & 1.01 & \textbf{0.93} \\
         &&MAPE& 320\% & 320\% & 448\% & 479\% & 330\% & 294\% & \textbf{254\%} \\
        \hline
    \end{tabular}}
    \caption{\label{tab:resultECG} Performance comparison on ECoG signals forecast.}
    \vspace{-0.5cm}
\end{table*}

The distance function $\text{dist}(v_i, v_j)$ in (\ref{eq:adjacency}) represents the road network distance from sensor $v_i$ to sensor $v_j$, producing an asymmetric adjacency matrix for a directed graph. Therefore, the symmetrized matrix $\hat{\mA} \coloneqq \frac{1}{2} (\mA + \mA^T)$ is taken to compute the wavelet basis matrix $\mW$ following Sec.~\ref{sec:multiresolution-analysis}.

Table \ref{tab:result_acc} shows the evaluation of different approaches on the two traffic datasets, while Table \ref{tab:trainTime} reports the training time per epoch of FTWGNN and DCRNN. Overall, although FTWGNN performs better than DCRNN by about only 10\%, it is significantly faster by about 5 times on average.


\subsection{Brain networks} \label{sec:brain}

\textit{Annotated Joints in Long-term Electrocorticography for 12 human participants} (AJILE12), publicly available at~\cite{brainnetdata}, records intracranial neural activity via the invasive ECoG, which involves implanting electrodes directly under the skull~\cite{ajile12}. For each participant, ECoG recordings are sporadically sampled at 500Hz in $7.4\pm2.2$ days (mean$\pm$std) from at least 64 electrodes, each of which is encoded with an unique set of Montreal Neurological Institute (MNI) x, y, z coordinates.

The proposed model is tested on the first one hour of recordings of subject number 5 with 116 good-quality electrodes. Subject 5 was chosen because he/she has the highest number of validated electrodes. Signals are downsampled to 1Hz, thus producing a network of 116 nodes, each with $3{,}600$ data points. Furthermore, the signals are augmented by applying the spline interpolation to get the upper and lower envelopes along with an average curve~\cite{melia2014filtering} (see Figure~\ref{fig:augmentedECG}). The adjacency matrix $\mA$ is obtained by solving task~(\ref{manifold.identify}), then the wavelet basis matrix $\mW$ is constructed based on Sec.~\ref{sec:multiresolution-analysis}.

Table \ref{tab:resultECG} reports the performance of different methods on the AJILE12 dataset for 1-, 5-, and 15-second prediction. Generally, errors are much higher than those in the traffic forecasting problem, since the connections within the brain network are much more complicated and ambiguous~\cite{breakspear2017dynamic}. High errors using HA and VAR methods show that the AJILE12 data follows no particular pattern or periodicity, making long-step prediction extremely challenging. Despite having a decent performance quantitatively, Figure~\ref{fig:expECG1s} demonstrates the superior performance of FTWGNN, in which DCRNN fails to approximate the trend and the magnitude of the signals. Even though FTWGNN performs well at 1-second prediction, it produces unstable and erroneous forecast at longer steps of 5 or 15 seconds. Meanwhile, similar to traffic prediction case, FTWGNN also sees a remarkable improvement in computation time by around 2 times on average (see Table \ref{tab:trainTime}).

\section{Conclusion} \label{sec:conclusion}

We propose a new class of spatial-temporal graph neural networks based on the theories of multiresolution analysis and wavelet theory on discrete spaces with RNN backbone, coined \textit{Fast Temporal Wavelet Graph Neural Network} (FTWGNN). Fundamentally, we employ \textit{Multiresolution Matrix Factorization} to factorize the underlying graph structure and extract its corresponding sparse wavelet basis that consequentially allows us to construct efficient wavelet transform and convolution on graph.  Experiments on real-world large-scale datasets show promising results and computational efficiency of FTGWNN in network time series modeling including traffic prediction and brain networks. Several future directions are:
\begin{enumerate*}[label= \textbf{(\roman*)}]
    \item investigating synchronization phenomena in brain networks \cite{honda2018mathematical};
    \item developing a robust model against outliers/missing data that appear frequently in practice.
\end{enumerate*}

\bibliography{paper}
\bibliographystyle{abbrvnat}


\appendix
\section{Multiresolution Matrix Factorization} \label{sec:mmf-appendix}

\subsection{Formal definitions}

\begin{definition} \label{def:rotation-matrix}
We say that $\mU \in \mathbb{R}^{n \times n}$ is an \textbf{elementary rotation of order $k$} (also called as a $k$-point rotation) if it is an orthogonal matrix of the form
$$\mU = \mI_{n - k} \oplus_{(i_1, \cdots, i_k)} \mO$$
for some $\sI = \{i_1, \cdots, i_k\} \subseteq [n]$ and $\mO \in \sS\sO(k)$. We denote the set of all such matrices as $\sS\sO_k(n)$.
\end{definition}

\begin{definition} \label{def:core-diagonal}
Given a set $\sS \subseteq [n]$, we say that a matrix $\mH \in \mathbb{R}^{n \times n}$ is $\sS$-core-diagonal if $\mH_{i, j} = 0$ unless $i, j \in \sS$ or $i = j$. Equivalently, $\mH$ is $\sS$-core-diagonal if it can be written in the form $\mH = \mD \oplus_{\sS} \overline{\mH}$, for some $\overline{H} \in \mathbb{R}^{\lvert \sS \rvert \times \lvert \sS \rvert}$ and $\mD$ is diagonal. We denote the set of all $\sS$-core-diagonal symmetric matrices of dimension $n$ as $\sH^{\sS}_n$.
\end{definition}

\begin{definition} \label{def:mmf}
Given an appropriate subset $\sO$ of the group $\sS\sO(n)$ of $n$-dimensional rotation matrices, a depth parameter $L \in \mathbb{N}$, and a sequence of integers $n = d_0 \ge d_1 \ge d_2 \ge \dots \ge d_L \ge 1$, a \textbf{Multiresolution Matrix Factorization (MMF)} of a symmetric matrix $\mA \in \mathbb{R}^{n \times n}$ over $\sO$ is a factorization of the form
\begin{equation} \label{eq:mmf}
\mA = \mU_1^T \mU_2^T \dots \mU_L^T \mH \mU_L \dots \mU_2 \mU_1,
\end{equation}
where each $\mU_\ell \in \sO$ satisfies $[\mU_\ell]_{[n] \setminus \sS_{\ell - 1}, [n] \setminus \sS_{\ell - 1}} = \mI_{n - d_\ell}$ for some nested sequence of sets $\sS_L \subseteq \cdots \subseteq \sS_1 \subseteq \sS_0 = [n]$ with $\lvert \sS_\ell \rvert = d_\ell$, and $\mH \in \sH^{\sS_L}_n$ is an $\sS_L$-core-diagonal matrix.
\end{definition}

\begin{definition} \label{def:factorizable}
We say that a symmetric matrix $\mA \in \mathbb{R}^{n \times n}$ is \textbf{fully multiresolution factorizable} over $\sO \subset \sS\sO(n)$ with $(d_1, \dots, d_L)$ if it has a decomposition of the form described in Def.~\ref{def:mmf}.
\end{definition}

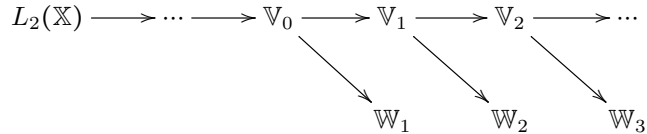
\begin{figure}[b]
$$
\xymatrix{
L_2(\sX) \ar[r] & \cdots \ar[r] & \sV_0 \ar[r] \ar[dr] & \sV_1 \ar[r] \ar[dr] & \sV_2 \ar[r] \ar[dr] & \cdots \\
& & & \sW_1 & \sW_2 & \sW_3
}
$$
\caption{\label{fig:subspaces}
Multiresolution analysis splits each function space $\sV_0, \sV_1, \dots$ into the direct sum of a smoother part $\sV_{\ell + 1}$ and a rougher part $\sW_{\ell + 1}$.
}
\end{figure}

\subsection{MMF optimization problem}

Finding the best MMF factorization to a symmetric matrix $\mA$ involves solving
\begin{equation}
\min_{\substack{\sS_L \subseteq \cdots \subseteq \sS_1 \subseteq \sS_0 = [n] \\ \mH \in \sH^{\sS_L}_n; \mU_1, \dots, \mU_L \in \sO}} \| \mA - \mU_1^T \dots \mU_L^T \mH \mU_L \dots \mU_1 \|.
\label{eq:mmf-opt}
\end{equation}
Assuming that we measure error in the Frobenius norm, (\ref{eq:mmf-opt}) is equivalent to
\begin{equation}
\min_{\substack{\sS_L \subseteq \cdots \subseteq \sS_1 \subseteq \sS_0 = [n] \\ \mU_1, \dots, \mU_L \in \sO}} \| \mU_L \dots \mU_1 \mA \mU_1^T \dots \mU_L^T \|^2_{\text{resi}},
\label{eq:mmf-resi}
\end{equation}
where $\| \cdot \|_{\text{resi}}^2$ is the squared residual norm 
$\|\mH \|_{\text{resi}}^2 = \sum_{i \neq j; (i, j) \not\in \sS_L \times \sS_L} \lvert \mH_{i, j} \rvert^2$. The optimization problem in (\ref{eq:mmf-opt}) and (\ref{eq:mmf-resi}) is equivalent to the following 2-level one:
\begin{equation}
\min_{\sS_L \subseteq \cdots \subseteq \sS_1 \subseteq \sS_0 = [n]} \min_{\mU_1, \dots, \mU_L \in \sO} \| \mU_L \dots \mU_1 \mA \mU_1^T \dots \mU_L^T \|^2_{\text{resi}}.
\label{eq:mmf-two-phases}
\end{equation}
There are two fundamental problems in solving this 2-level optimization:
\begin{compactitem}
\item For the inner optimization, the variables (i.e. Givens rotations $\mU_1, \dots, \mU_L$) must satisfy the orthogonality constraints.
\item For the outer optimization, finding the optimal nested sequence of indices $\sS_L \subseteq \cdots \subseteq \sS_1 \subseteq \sS_0 = [n]$ is a combinatorics problem, given an exponential search space.
\end{compactitem}
In order to address these above problems, \cite{pmlr-v196-hy22a} proposes a learning algorithm combining Stiefel manifold optimization and Reinforcement Learning (RL) for the inner and outer optimization, respectively. In this paper, we assume that a nested sequence of indices $\sS_L \subseteq \cdots \subseteq \sS_1 \subseteq \sS_0 = [n]$ is given by a fast heuristics instead of computationally expensive RL. There are several heuristics to find the nested sequence, for example: clustering based on similarity between rows \cite{pmlr-v32-kondor14} \cite{DBLP:journals/corr/KondorTM15}. In section \ref{sec:stiefel}, we introduce the solution for the inner problem.

\subsection{Stiefel manifold optimization} \label{sec:stiefel}

In order to solve the inner optimization problem of (\ref{eq:mmf-two-phases}), 
we consider the following generic optimization with orthogonality constraints:
\begin{equation}
\min_{\mX \in \mathbb{R}^{n \times p}} \mathcal{F}(\mX), \ \ \text{s.t.} \ \ \mX^T \mX = \mI_p,
\label{eq:opt-prob}
\end{equation}
where $\mI_p$ is the identity matrix and $\mathcal{F}(\mX): \mathbb{R}^{n \times p} \rightarrow \mathbb{R}$ is a differentiable function. The feasible set $\mathcal{V}_p(\mathbb{R}^n) = \{\mX \in \mathbb{R}^{n \times p}: \mX^T \mX = \mI_p\}$ is referred to as the Stiefel manifold of $p$ orthonormal vectors in $\mathbb{R}^{n}$. We will view $\mathcal{V}_p(\mathbb{R}^n)$ as an embedded submanifold of $\mathbb{R}^{n \times p}$. In the case there are more than one orthogonal constraints, (\ref{eq:opt-prob}) is written as
\begin{equation}
\min_{\mX_1 \in \mathcal{V}_{p_1}(\mathbb{R}^{n_1}), \dots, \mX_q \in \mathcal{V}_{p_q}(\mathbb{R}^{n_q})} \mathcal{F}(\mX_1, \dots, \mX_q)
\label{eq:opt-prob-extended}
\end{equation}
where there are $q$ variables with corresponding $q$ orthogonal constraints. 
In the MMF optimization problem (\ref{eq:mmf-two-phases}), suppose we are already given $\sS_L \subseteq \cdots \subseteq \sS_1 \subseteq \sS_0 = [n]$ meaning that the indices of active rows/columns at each resolution were already determined, for simplicity. In this case, we have $q = L$ number of variables such that each variable $\mX_\ell = \mO_\ell \in \mathbb{R}^{k \times k}$, where $\mU_\ell = \mI_{n - k} \oplus_{\sI_\ell} \mO_\ell \in \mathbb{R}^{n \times n}$ in which $\sI_\ell$ is a subset of $k$ indices from $\sS_\ell$, must satisfy the orthogonality constraint. The corresponding objective function is 
\begin{equation}
\mathcal{F}(\mO_1, \dots, \mO_L) = \| \mU_L \dots \mU_1 \mA \mU_1^T \dots \mU_L^T \|^2_{\text{resi}}.
\label{eq:mmf-core}
\end{equation}
Therefore, we can cast the inner problem of (\ref{eq:mmf-two-phases}) as an optimization problem on the Stiefel manifold, and solve it by the specialized steepest gradient descent \cite{Tagare2011NotesOO}.

\begin{figure}
  \centering
    \includegraphics[width=0.5\textwidth]{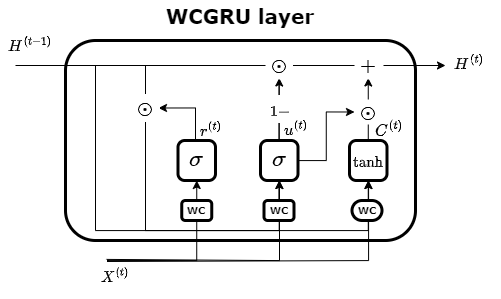}
    \centering
    \caption{Architecture for the Wavelet Convolutional Gated Recurrent Unit. \textbf{WC}: graph wavelet convolution given MMF's wavelet basis}
  \label{fig:architecture2}    
\end{figure}

\section{Baseline settings} \label{app:baseline}

\textbf{HA} \quad Historical average, which models traffic flow as a seasonal / periodic process, and uses the weighted average of previous seasons / periods as a prediction.

$\textbf{ARIMA}_{kal}$ \quad Auto-Regressive Integrated Moving Average Model with Kalman Filter, implemented by the \textit{statsmodel} package in Python.

\textbf{VAR} \quad Vector Auto-regressive model~\cite{hamilton2020time} with orders $(3, 0, 1)$, implemented by the Python \textit{statsmodel} package.

\textbf{LR} \quad Linear regression with 5 historical observations.

\textbf{SVR} \quad Linear Support Vector Regression~\cite{smola2004svr} with 5 historical observations. 

\textbf{FNN} \quad Feed forward neural network with two hidden layers, each with 256 units. The initial learning rate is $1e^{-3}$, and the decay rate is $1e^{-1}$ per 20 epochs. In addition, for all hidden layers, dropout with ratio 0.5 and L2 weight decay $1e^{-2}$ is used. The model is trained to minimize the MAE with batch size 64.

\textbf{FC-LSTM} \quad The encoder-decoder framework using LSTM with peephole~\cite{sutskever2014sequence}. The encoder and the decoder contain two recurrent layers, each of which consists of 256 LSTM units, with an L1 weight decay rate $2e^{-5}$ and an L2 weight decay rate $5e^{-4}$. The initial learning rate is $1e^{-4}$ and the decay rate is $1e^{-1}$ per 20 epochs.

\textbf{DCRNN}~\cite{dcrnn}, \textbf{GWaveNet}~\cite{wu2019graph}, \textbf{STGCN}~\cite{han2020stgcn} \quad Settings follow their original work.

\section{FTWGNN settings} \label{app:ftwgnn}
\textbf{Data preparation} \quad For all datasets, the train/validation/test ratio is $0.7/0.2/0.1$, divided into batch size 64.

\textbf{Adjacency matrix} \quad The $k$-neighborhood of the traffic network in Eq.~\ref{eq:adjacency} is thresholded by $k=0.01$, while for the brain network, the parameter $\lambda_A$ is set to $1e^{-5}$ in Task (\ref{manifold.identify}).

\textbf{Wavelet basis} \quad For the traffic datasets, 100 mother wavelets are extracted, \ie, $L=100$, while for the AJILE12 dataset, $L=10$ was used.

\textbf{Model architecture} \quad For the wavelet convolution RNN, both encoder and decoder contains two recurrent layers, each with 64 units. The initial learning rate is $1e^{-2}$, decaying by $\frac{1}{10}$ per 20 epochs; the dropout ratio is $0.1$; and the maximum diffusion step, \ie, $K$, is set to 2. In addition, the optimizer is the Adam optimizer~\cite{kingma2014adam}.


    

\section{Vizualization of the AJILE12 data}

\begin{figure}[h]
  \centering
    \includegraphics[width=0.8\textwidth]{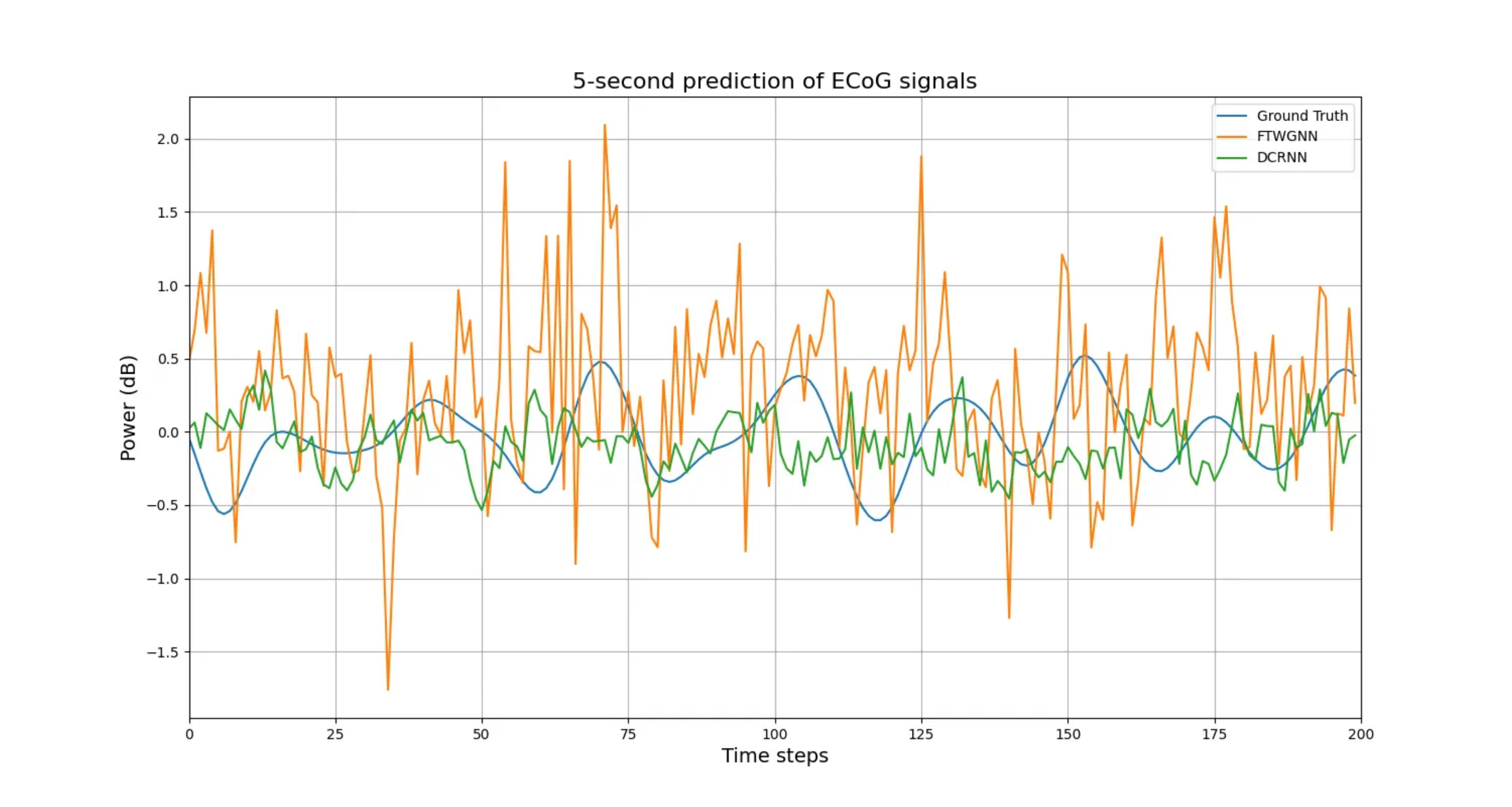}
    \caption{5-second prediction of ECoG signals.}
  \label{fig:expECG5s}    
\end{figure}

\begin{figure*}[!ht]
    \subfloat[$\ell=1$]
    {\includegraphics[width=0.33\textwidth]{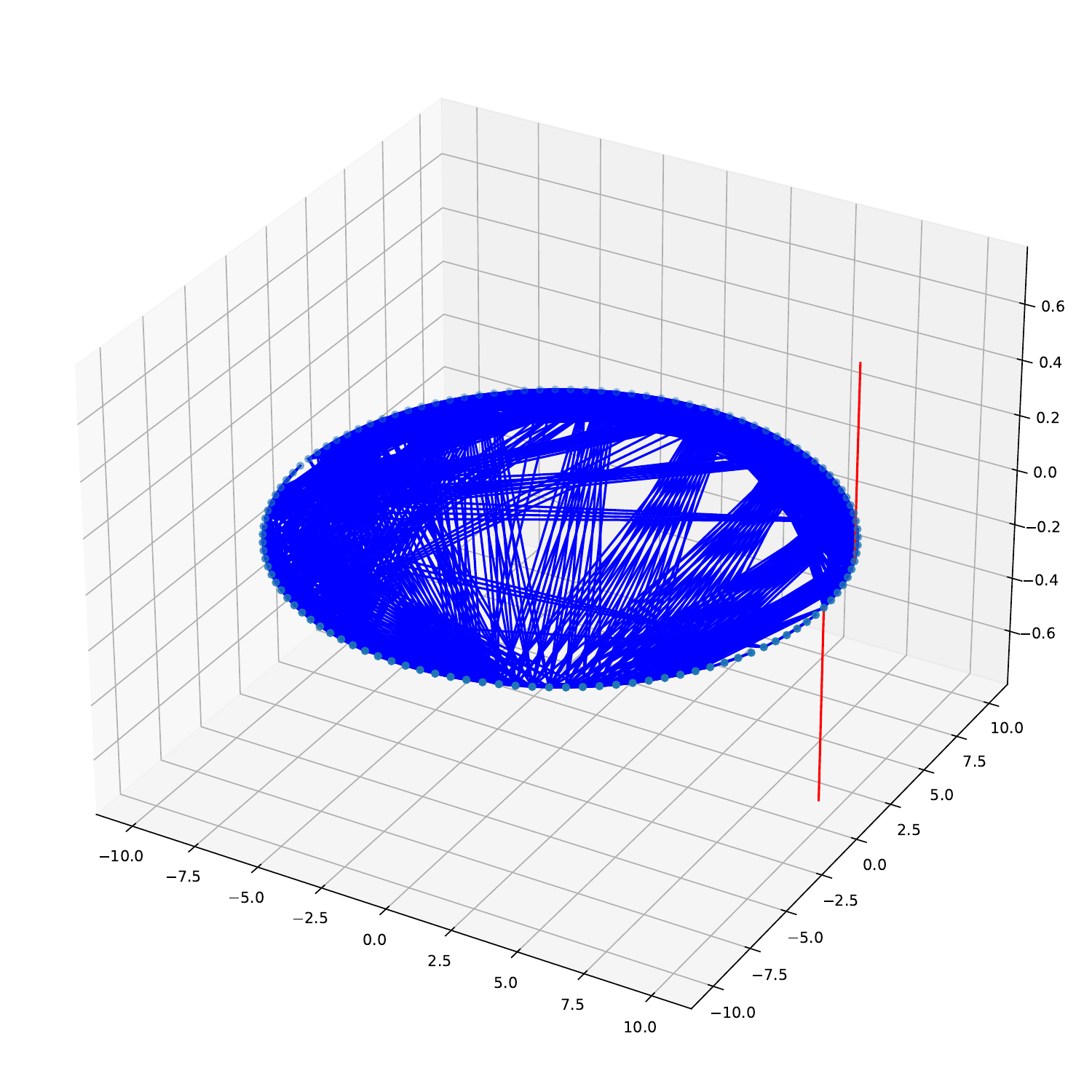}}
    \subfloat[$\ell=20$]
    {\includegraphics[width=0.33\textwidth]{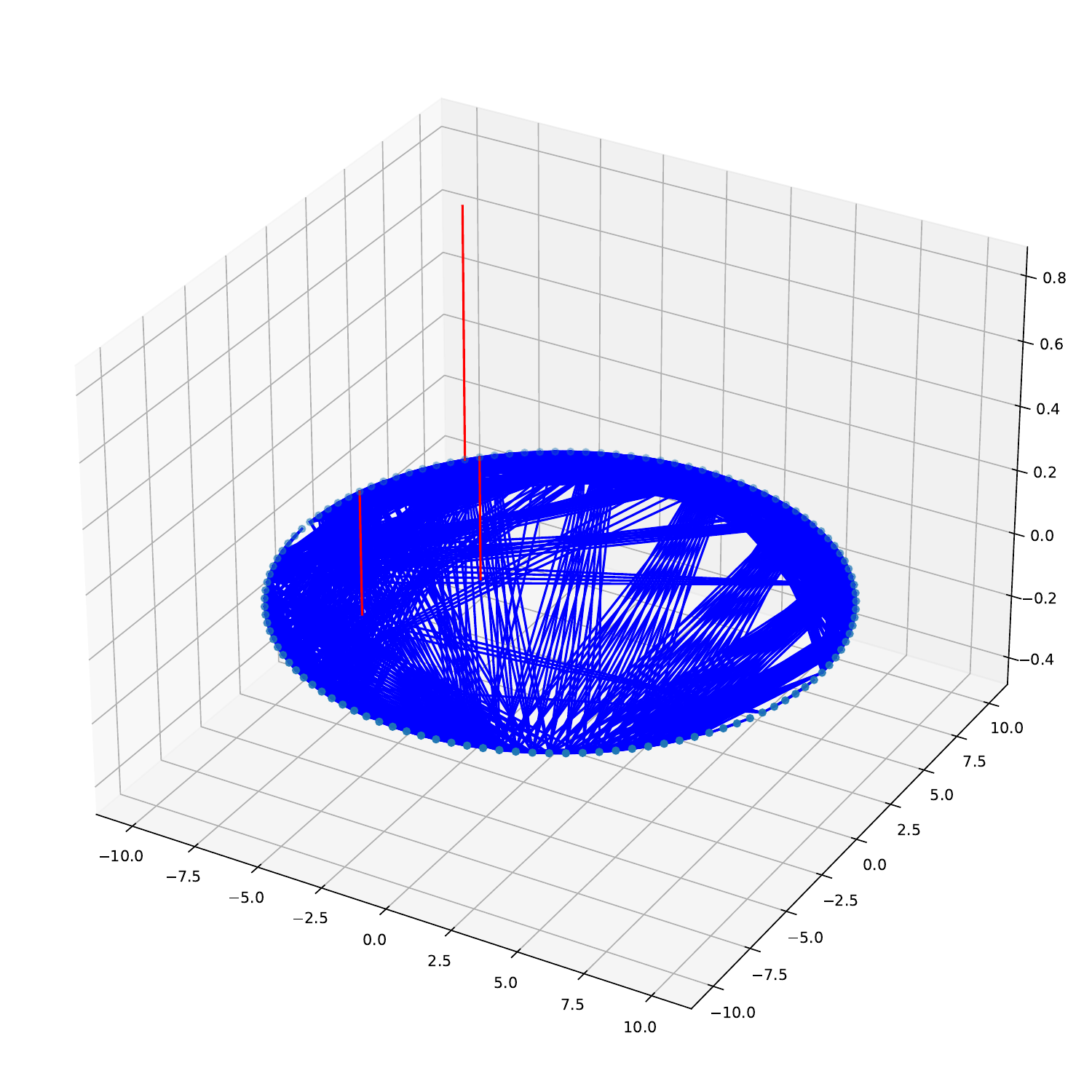}}
    \subfloat[$\ell=49$]
    {\includegraphics[width=0.33\textwidth]{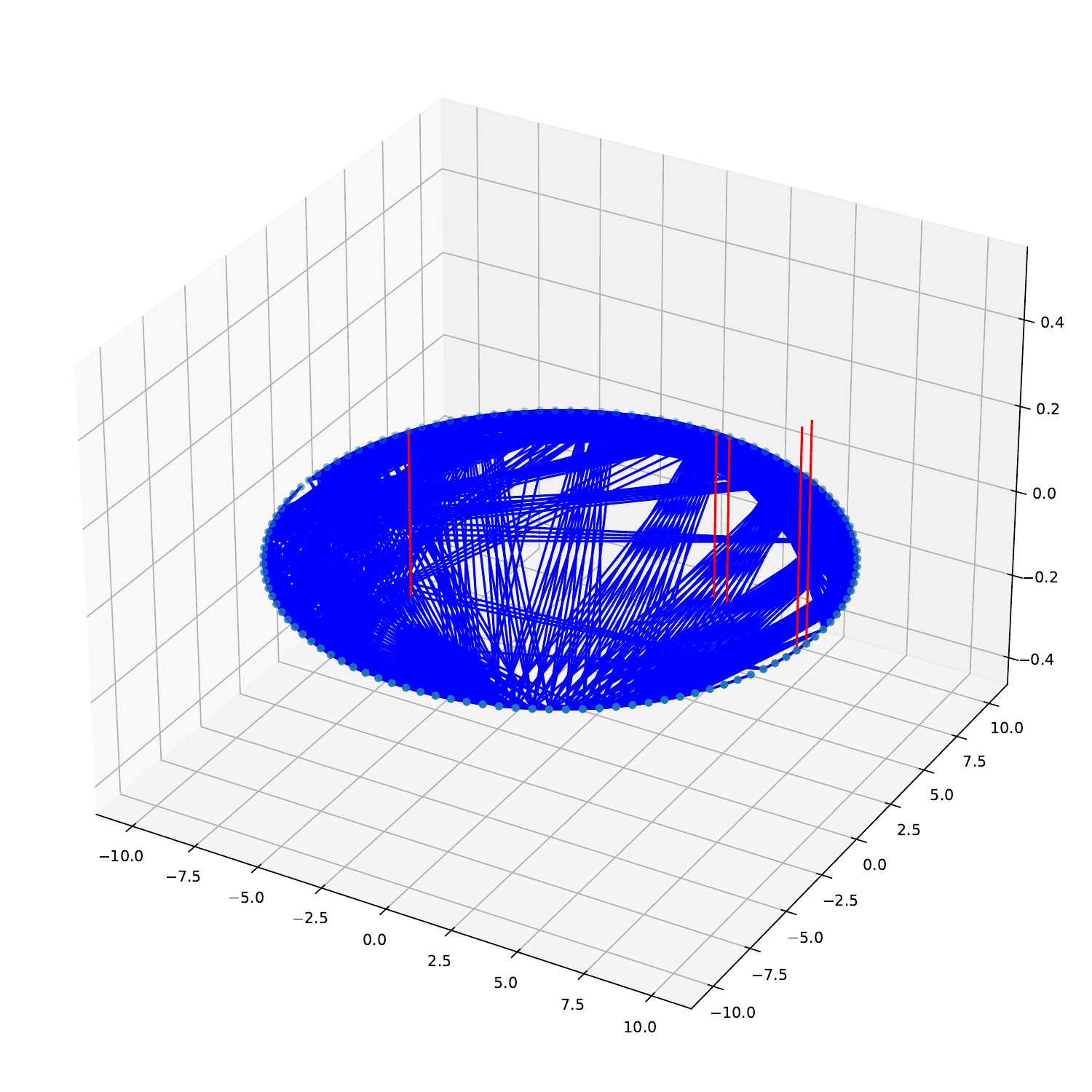}}
    
    
    \caption{Visualization of some of the wavelets on the brain network of 116 nodes. The low index wavelets (low $\ell$) are highly localized, whereas the high index ones are smoother and more dispersed.}
    \label{fig:waveletVisual}
\end{figure*}


\end{document}